\documentclass{article} 
\usepackage{iclr2022_conference,times}

\usepackage{amsmath,amsfonts,bm}

\def\eqref#1{equation~\ref{#1}}

\def\1{\bm{1}}

\DeclareMathAlphabet{\mathsfit}{\encodingdefault}{\sfdefault}{m}{sl}
\SetMathAlphabet{\mathsfit}{bold}{\encodingdefault}{\sfdefault}{bx}{n}

\usepackage{hyperref}
\usepackage{url}
\usepackage{booktabs}       
\usepackage{nicefrac}       
\usepackage{microtype}      
\usepackage{multirow}
\usepackage{amsmath}
\usepackage{amsfonts}
\usepackage{amssymb}
\usepackage{graphicx}
\usepackage{subfigure} 
\usepackage{wrapfig}
\usepackage{caption} 
\usepackage{subfigure}
\usepackage{color}
\usepackage{algorithm}  
\usepackage{algpseudocode}  
 \usepackage{threeparttable}
\usepackage{xcolor}
\newcommand{\red}[1]{\textcolor{red}{#1}}

\title{DeepFIB: Self-Imputation for Time Series \\ Anomaly Detection}

\author{ Minhao Liu, Zhijian Xu, Qiang Xu \\
\texttt{\{mhliu,  zjxu,  qxu\}@cse.cuhk.edu.hk} \\
}

\iclrfinalcopy 
\begin{document}

\maketitle



\begin{abstract}
Time series (TS) anomaly detection (AD) plays an essential role in various applications, e.g., fraud detection in finance and healthcare monitoring.
Due to the inherently unpredictable and highly varied nature of anomalies and the lack of anomaly labels in historical data, the AD problem is typically formulated as an unsupervised learning problem. The performance of existing solutions is often not satisfactory, especially in data-scarce scenarios.
To tackle this problem, we propose a novel self-supervised learning technique for AD in time series, namely \emph{DeepFIB}. We model the problem as a \emph{Fill In the Blank} game by masking some elements in the TS and imputing them with the rest. 
Considering the two common anomaly shapes (point- or sequence-outliers) in TS data, we implement two masking strategies with many self-generated training samples. The corresponding self-imputation networks can extract more robust temporal relations than existing AD solutions and effectively facilitate identifying the two types of anomalies.  
For continuous outliers, we also propose an anomaly localization algorithm that dramatically reduces AD errors. Experiments on various real-world TS datasets demonstrate that DeepFIB outperforms state-of-the-art methods by a large margin, achieving up to $65.2\%$ relative improvement in F1-score.

\end{abstract}

\section{Introduction}


Anomaly detection (AD) in time series (TS) data has numerous applications across various domains. Examples include fault and damage detection in industry~\citep{Hundman2018DetectingSA}, intrusion detection in cybersecurity~\citep{Feng2021TimeSA}, and fraud detection in finance~\citep{Zheng2018GenerativeAN} or healthcare~\citep{Zhou2019BeatGANAR}, to name a few. 

Generally speaking, an anomaly/outlier is an observation that deviates considerably from some concept of normality~\citep{Ruff2021ADReview}. The somewhat “vague” definition itself tells the challenges of the AD problem arising from the rare and unpredictable nature of anomalies. With the lack of anomaly labels in historical data, most AD approaches try to learn the expected values of time-series data in an unsupervised manner~\citep{BlazquezGarcia2021ARO}. Various techniques use different means (e.g., distance-based methods~\citep{Angiulli2002FastOD}, predictive methods~\citep{Holt2004ForecastingSA, Yu2016AnIA,Deng2021GraphNN} or reconstruction-based methods~\citep{Shyu2003ANA, Malhotra2016LSTMbasedEF, Zhang2019ADN,Shen2021TimeSA}) to obtain this expected value, and then compute how far it is from the actual observation to decide whether or not it is an anomaly.

While existing solutions have shown superior performance on some time series AD tasks, they are still far from satisfactory. For example, for the six ECG datasets in~\citep{Keogh2005ECG}, the average F1-score of state-of-the-art solutions~\citep{Kieu2019OutlierDF,Shen2021TimeSA} with model ensembles are barely over $40\%$. Other than the TS data' complexity issues, one primary reason is that the available data is often scarce while deep learning algorithms are notoriously data-hungry. 

Recently, self-supervised learning (SSL) that enlarges the training dataset without manual labels has attracted lots of attention, and it has achieved great success in representation learning in computer vision~\citep{Zhang2016ColorfulIC,Pathak2016ContextEF,Chen2020ASF}, natural language processing~\citep{Devlin2019BERTPO}, and graph learning~\citep{Hu2020StrategiesFP} areas. There are also a few SSL techniques for time series analysis proposed in the literature. Most of them~\citep{Falck2020ContrastiveRL,Saeed2021FederatedSL,fan2020selfsupervised} craft contrastive TS examples for classification tasks. \citep{Deldari2021TimeSC} also leverages contrastive learning for change point detection in time series. 




\begin{wrapfigure}{r}{0.6\textwidth}
\centering
\subfigure[point-wise]{
\includegraphics[width=3.5cm]{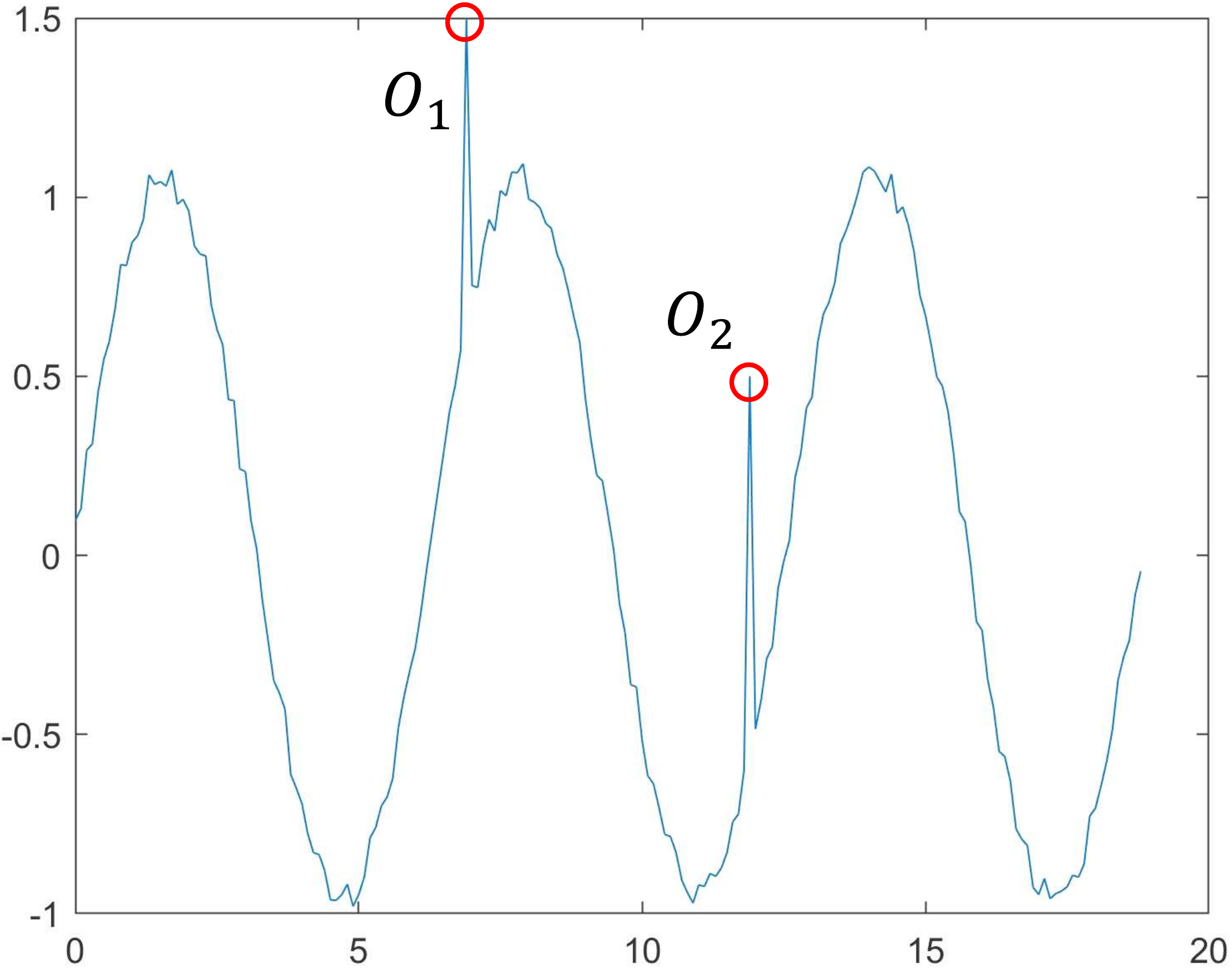}
}
\hspace{3mm}
\subfigure[sequence-wise]{
\includegraphics[width=3.5cm]{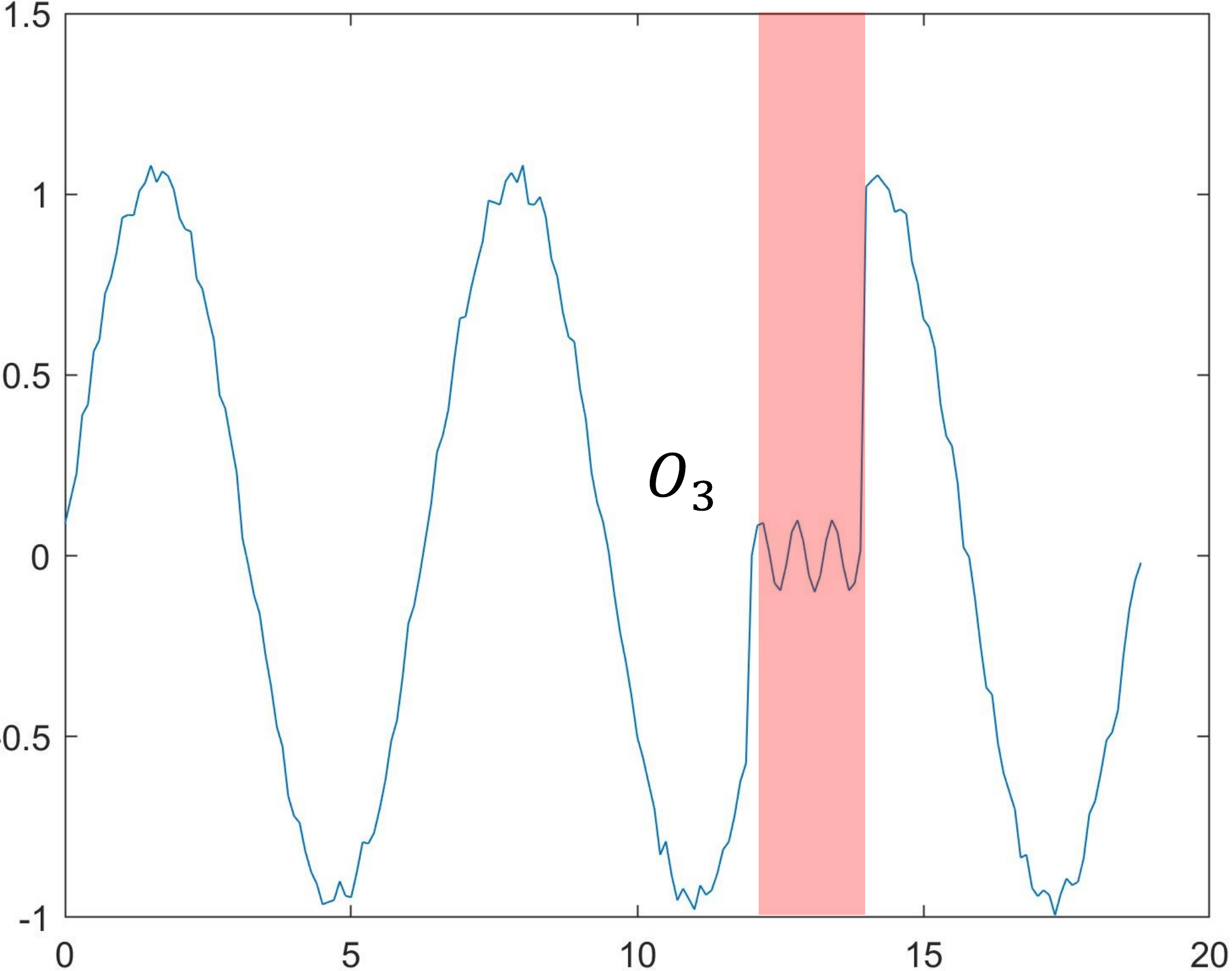}
}
\vspace{-5pt}
\caption{Anomalies in time series.}
\label{fig:types}
\vspace{-5pt}
\end{wrapfigure}

While interesting, the above SSL techniques do not apply to the AD task because detecting anomalies in time series requires fine-grained models at the element level. In this work, inspired by the context encoder for visual feature learning~\citep{Pathak2016ContextEF} and the BERT model for language representation learning~\citep{Devlin2019BERTPO}, we propose a novel self-supervised learning technique for time series anomaly detection, namely \emph{DeepFIB}. To be specific, we model the problem as a \emph{Fill In the Blank} game by masking some elements in the TS and imputing them with other elements. This is achieved by revising the TS forecasting model \emph{SCINet}~\citep{Liu2021TimeSI} for the TS imputation task, in which the masked elements are regarded as missing values for imputation. Such self-imputation strategies facilitate generating a large amount of training samples for temporal relation extraction. As anomalies in time series manifest themselves as either discrete points or subsequences (see Fig.~\ref{fig:types}), correspondingly, we propose two kinds of masking strategies and use them to generate two pre-trained models. They are biased towards recovering from \emph{point-wise} anomalies (\emph{DeepFIB-p} model for \emph{point outliers}) and \emph{sequence-wise} anomalies (\emph{DeepFIB-s} model for \emph{continuous outliers}), respectively. To the best of our knowledge, this is the first SSL work for time series anomaly detection. 

Generally speaking, AD solutions have difficulty detecting sequence-wise anomalies because it is hard to tell the real outliers from their neighboring normal elements due to their interplay. 
To tackle this problem, we propose a novel anomaly localization algorithm to locate the precise start and end positions of continuous outliers. As a post-processing step, we conduct a local search after determining the existence of sequence-wise anomalies within a timing window with our \emph{DeepFIB-s} model. By doing so, the detection accuracy for continuous outliers is significantly improved.

We conduct experiments on several commonly-used time series benchmarks, and results show that DeepFIB consistently outperforms state-of-the-art solutions. In particular, the average F1-score of DeepFIB for the six ECG datasets is more than $62\%$, achieving nearly $50\%$ relative improvement. 


\section{Related Work}\label{sec:2-relatedwork}

In this section, we mainly discuss recent deep learning-based time series AD approaches. A comprehensive survey on the traditional techniques can be found in~\citep{Gupta2014TSADSurvey}.



\begin{figure*}[h]
    \centering
    \includegraphics[width=0.9\linewidth]{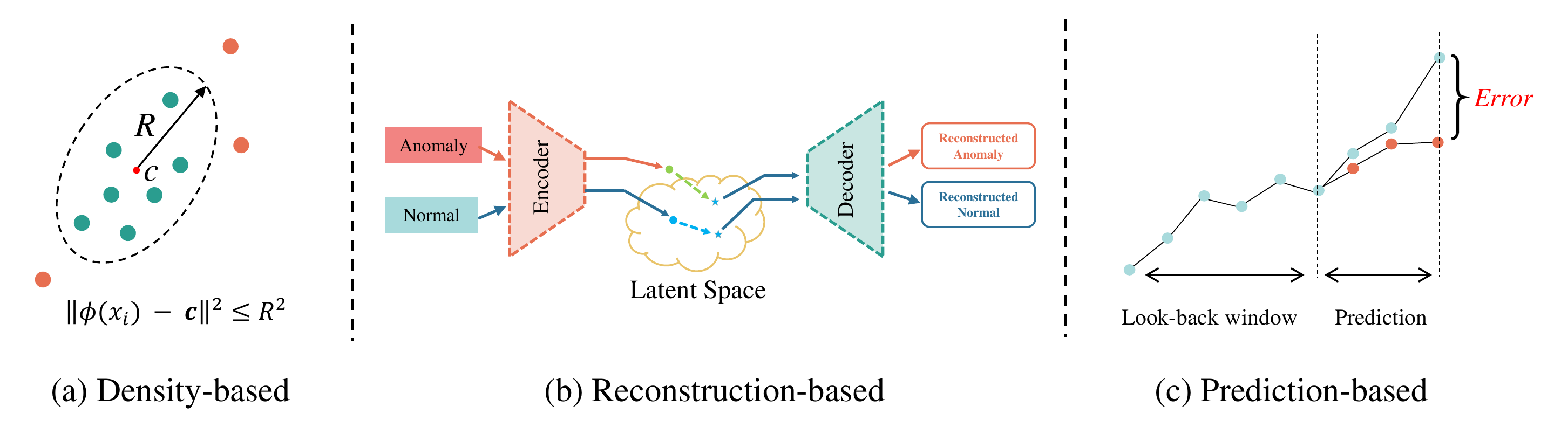}
    \caption{Existing time series anomaly detection architectures.
    }
    \label{fig:RW}
\vspace{-5pt}
\end{figure*}

Existing anomaly detection approaches can be broadly categorized into three types (see Fig.~\ref{fig:RW}): (i) \emph{Density-based} methods consider the normal instances compact in the latent space and identify anomalies with one-class classifiers or likelihood measurements~\citep{Su2019RobustAD,Shen2020TimeseriesAD,Feng2021TimeSA}.
(ii) \emph{Reconstruction-based} methods use recurrent auto-encoders~(RAE)~\citep{Malhotra2016LSTMbasedEF,Yoo2021RecurrentRN,Kieu2019OutlierDF,Shen2021TimeSA, Zhang2019ADN} or deep generative models such as recurrent VAEs~\citep{Park2018AMA} or GANs~\citep{Li2019MADGANMA,Zhou2019BeatGANAR} for reconstruction. The reconstruction errors are used as anomaly scores. 
(iii) \emph{Prediction-based} methods rely on predictive models~\citep{Bontemps2016CollectiveAD,Deng2021GraphNN,Chen2021LearningGS} 
and use the prediction errors as anomaly scores.


While the above methods have been successfully used in many real-world applications, practical AD tasks still have lots of room for improvement, especially in data-scarce scenarios. Unlike existing AD approaches, the proposed mask-and-impute method in \emph{DeepFIB} exploits the unique property of TS data that missing values can be effectively imputed~\citep{Fang2020TSImpute}. By constructing many training samples via self-imputation, \emph{DeepFIB} extracts robust temporal relations of TS data and improves AD accuracy dramatically. Moreover, for the more challenging sequence-wise anomalies, most prior work assumes a user-defined fixed-length for anomaly subsequences~\citep{Cook2020AnomalyDF} or simplifies the problem by stating all the continuous outliers have been correctly detected as long as one of the points is detected~\citep{Su2019RobustAD,Shen2020TimeseriesAD}. In \emph{DeepFIB}, we lift these assumptions and try to locate the exact location of sequence-wise anomalies.

 \section{Method}\label{sec:3-proposed solution}


In this section, we first introduce the overall self-imputation framework in DeepFIB and then discuss the separate AD models for detecting point- and sequence-wise anomalies with different mask-and-impute strategies, namely \emph{DeepFIB-p} and \emph{DeepFIB-s}, respectively. Next, we describe the TS imputation method used in \emph{DeepFIB}, based on an existing TS forecasting approach \emph{SCINet}~\citep{Liu2021TimeSI}. Finally, we present our anomaly localization algorithm for continuous outliers.

\subsection{Self-Imputation for Anomaly Detection}\label{sec:maskimputation}

Given a set of multivariate time series wherein $X_s = \left \{ x_1,x_2,...,x_{T_s} \right \} \epsilon \mathbb{R}^{d \times T_s} $ ($T_s$ is the length of the $s_{th}$ time series $X_s$), the objective of the AD task is to find all anomalous points $x_t \in \mathbb{R}^{d}$ ($d$ is the number of variates) and anomalous subsequences  $X_{t,\tau} =\left \{x_{t-\tau+1}, ...,x_{t} \right \}$.

The critical issue to solve the above problem is obtaining an expected value for each element in the TS, which requires a large amount of training data to learn from, especially for deep learning-based solutions. However, time-series data are often scarce, significantly restricting the effectiveness of learning-based AD solutions. 

DeepFIB is a simple yet effective SSL technique to tackle the above problem. We model this problem as a \emph{Fill In the Blank} game by randomly masking some elements in the TS and imputing them with the rest. Such self-imputation strategies generate many training samples from every time series and hence dramatically improve temporal learning capabilities. 

\vspace{-5pt}
\begin{figure}[h]
    \centering
    \includegraphics[width=0.9\linewidth]{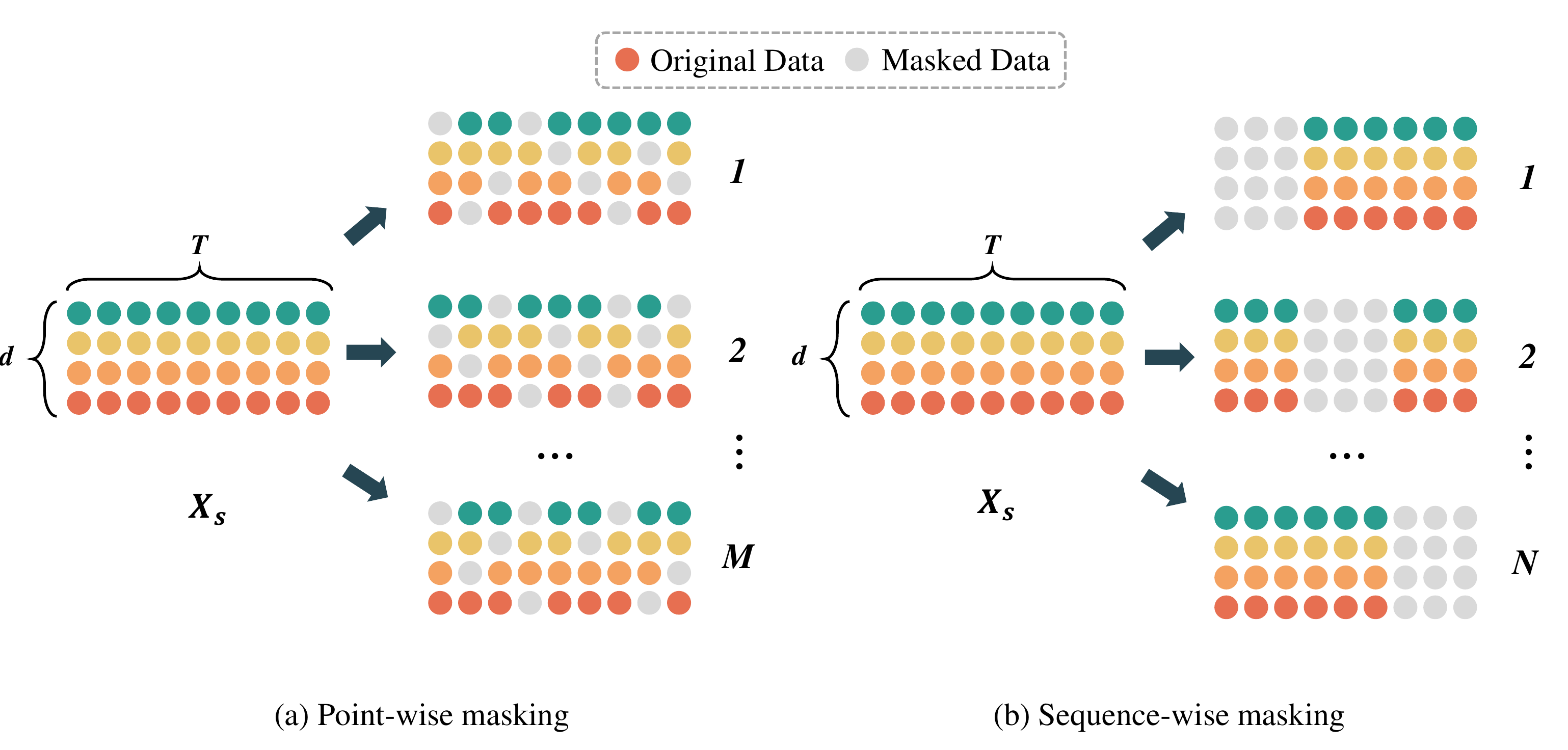}
    \caption{Self-imputation strategies of \emph{DeepFIB-p} and \emph{DeepFIB-s}.
    }
    \label{fig:Masking}
    \vspace{-5pt}
\end{figure}

In particular, we propose to train two self-imputation models~(Fig.~\ref{fig:Masking}), biased towards point- and sequence-wise anomalies in the TS data, respectively. 


\begin{itemize}
    \item \emph{DeepFIB-p model} targets point outliers, as shown in Fig.~\ref{fig:Masking}(a), in which we mask discrete elements and rely on the local temporal relations extracted from neighboring elements for reconstruction. 
    For each time series $X_s$, we generate $M$ training samples by masking it $M$ times with randomly-selected yet \emph{non-overlapping} $\frac{d \times T_s}{M}$ elements. 

    \item \emph{DeepFIB-s model} targets continuous outliers, as shown in Fig.~\ref{fig:Masking}(b), in which we mask continuous elements and rely on predictive models for reconstruction. 
    For each time series $X_s$, we evenly divide it into $N$ \emph{non-overlapping} sub-sequences as $  \left \{X^{d \times {\frac{T_s}{N}}}_{s,i}, i \in \left [0,N-1  \right ]  \right \} $ and generate $N$ training samples by masking one of them each time. 
\end{itemize}






During training, for each time series $X_s$, we obtain a set of non-overlapped imputed data with the above model and integrate them together results in a reconstructed time series $\widehat{X_s}$ (i.e., $\widehat{X_s}$\emph{-p} for \emph{DeepFIB-p} model and $\widehat{X_s}$\emph{-s} for \emph{DeepFIB-s} model).
The training loss for both models are defined as the reconstruction errors between the input time series and the reconstructed one:
$$
\mathcal{L} = \frac{1}{T_s}\sum_{t=1}^{T_s}\left \| x_t - \widehat{x_t} \right \| \eqno{(1)}
$$

where $x_t$ is the original input value at time step $t$ and the $\widehat{x_t}$ denotes the reconstructed value from the corresponding model, and $\left \| \cdot  \right \|$ is the L1-norm of a vector.

During testing, to detect point outliers with the \emph{DeepFIB-p} model, we simply use the residual error as the anomaly score, defined as $e_t = \sum_{i=0}^{d}\left | \widehat{x_{t}}^i-x_{t}^i \right |$, and when $e_t$ is larger than a threshold value $\lambda_p$, time step $t$ is regarded as an outlier. In contrast, for continuous outliers, we use dynamic time warping (DTW)~\citep{Sakoe1978DynamicPA} distance metrics as our anomaly scoring mechanism, which measures the similarity between the input time series $X$ and reconstructed sequence $\widehat{X}$. If $DTW(X, \widehat{X})$ is above a threshold value $\lambda_s$, a sequence-wise anomaly is detected.

\subsection{Time Series Imputation in DeepFIB}

While the time-series data imputation problem has been investigated for decades~\citep{Fang2020TSImpute}, there are still lots of rooms for improvement and various deep learning models are proposed recently~\citep{Cao2018BRITSBR,Liu2019NAOMINM,Luo2019EGANEG}.  

\begin{wrapfigure}{r}{0.5\textwidth} 
    \centering
    \includegraphics[width=0.8\linewidth]{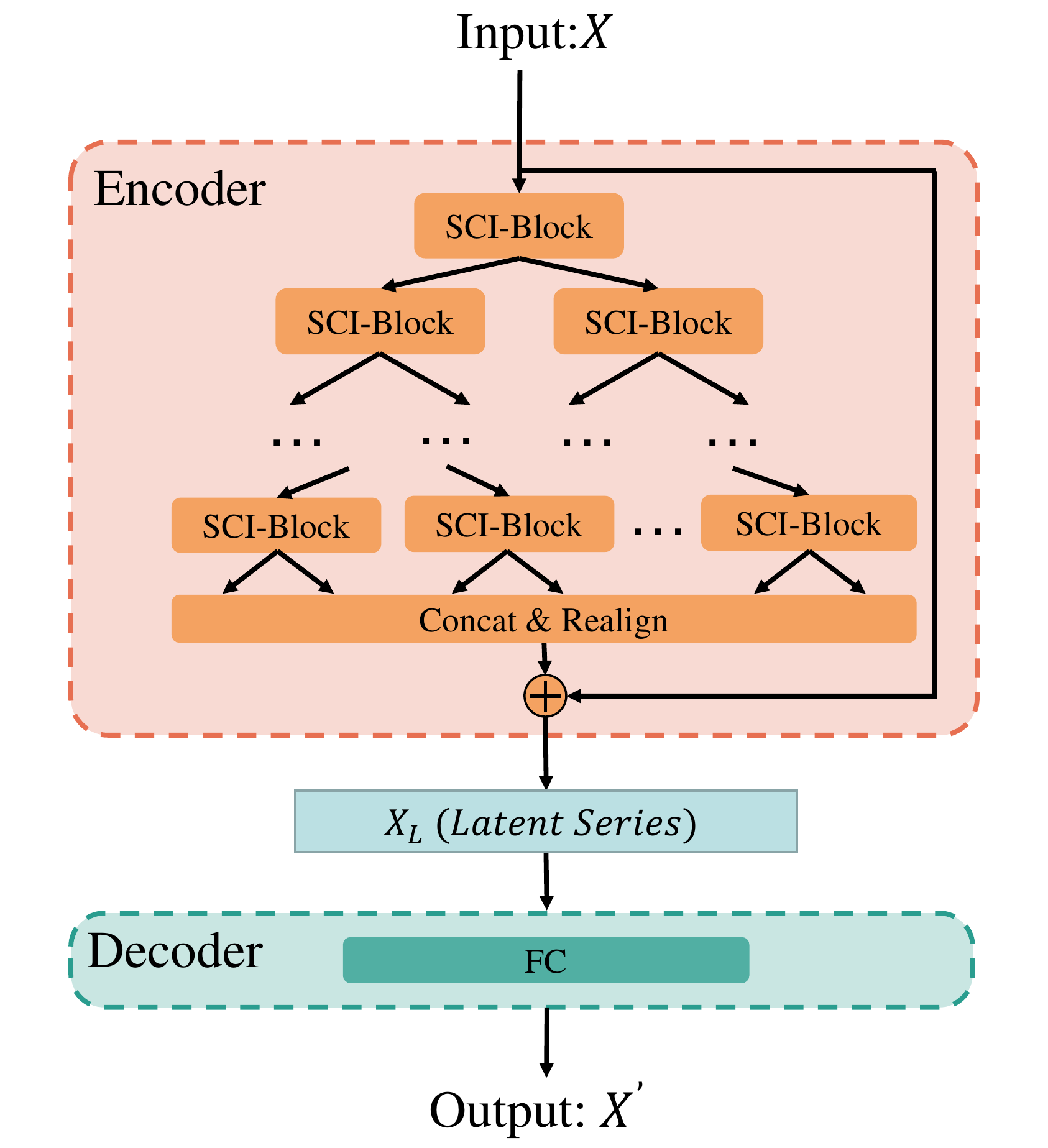}
   \caption{The structure of the SCINet.}
    \label{fig:SCINet}
    \vspace{-5pt}
\end{wrapfigure}

SCINet~\citep{Liu2021TimeSI}(Fig.~\ref{fig:SCINet}) is an encoder-decoder architecture motivated by the unique characteristics of time series data. It incorporates a series of SCI-Blocks that conduct down-sampled convolutions and interactive learning to capture temporal features at various resolutions and effectively blend them in a hierarchical manner. Considering the highly-effective temporal relation extraction capability of SCINet when compared to other sequence models, we propose to revise it for the TS imputation task.
More details about \emph{SCINet} can be found in~\citep{Liu2021TimeSI}.

\begin{figure}[h]
    \centering
    \includegraphics[width=0.95\linewidth]{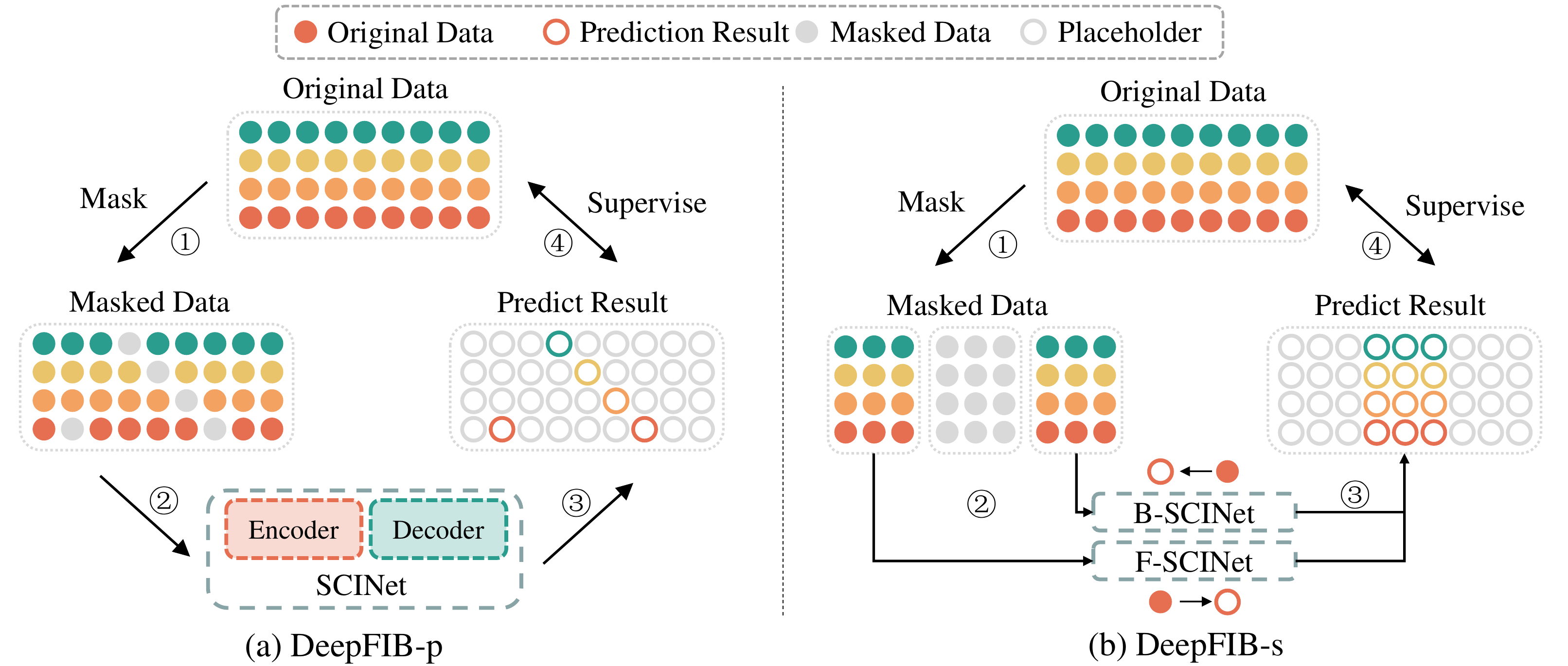}
    \caption{Time series imputation in DeepFIB.
    }
    \label{fig:imputation}
    \vspace{-5pt}
\end{figure}

To impute the missing elements from the two masking strategies with \emph{DeepFIB-p} and \emph{DeepFIB-s} models, 
we simply change the supervisions for the decoder part accordingly. 
For point imputation, we use the original input sequence as the supervision of our \emph{DeepFIB-p} model, making it a reconstruction structure. By doing so, the model concentrates more on the local temporal relations inside the timing window for imputing discrete missing data, as shown in  Fig.~\ref{fig:imputation}(a).
As for continuous imputation, we propose to change SCINet as a bidirectional forecasting structure in our \emph{DeepFIB-s} model, with the masked sub-sequence as supervision. As shown in Fig.~\ref{fig:imputation}(b), the two sub-models, namely \emph{F-SCINet} and \emph{B-SCINet}, are used to conduct forecasting in the forward and backward directions, respectively. 
By doing so, the model can aggregate the temporal features from both directions and learn a robust long-term temporal relations for imputing continuous missing data. 


\subsection{Anomaly Localization Algorithm}

During inference, we use a sliding window with stride $\mu$ to walk through the time series and find anomalies in each window. For sequence-wise anomalies, without knowing their positions a priori, we could mask some normal elements in the window and use those unmasked outliers for prediction (see Fig.~\ref{fig:imputation}(b)), thereby leading to mispredictions. 
To tackle this problem, we propose to conduct a local search for the precise locations of the sequence-wise anomalies.

\begin{figure}[h]
    \centering
    \includegraphics[width=0.75\linewidth]{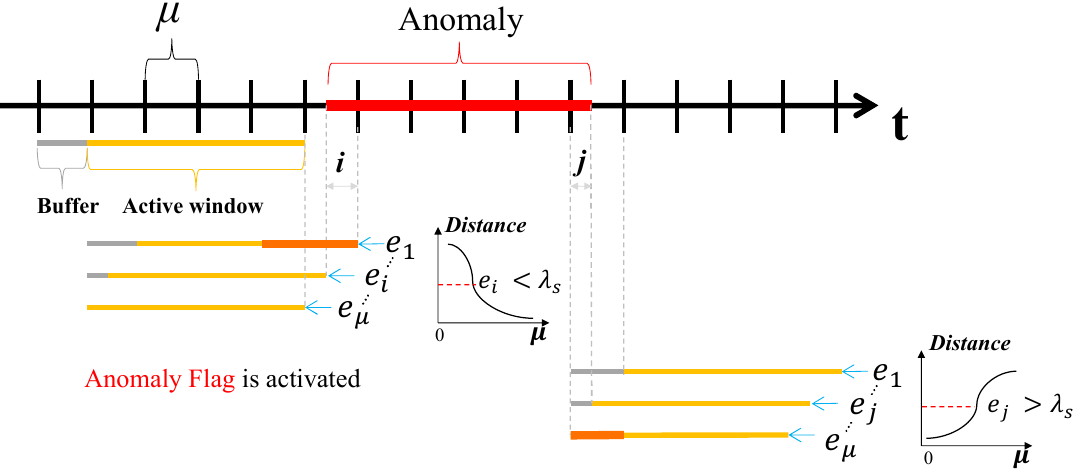}
    \caption{Anomaly localization algorithm. 
    }
    \label{fig:Srepir}
\end{figure}



As shown in Fig.~\ref{fig:Srepir}, 
the \emph{Active window} are the current input sequence to the DeepFIB-s model with length $\omega$ ($\omega > \mu$), i.e., $X_t = \left \{ x_t,x_{t+1},...,x_{t+\omega-1} \right \}$ at time step $t$. When the \emph{DTW} distance between the original time series in the \emph{Active window} and the imputed sequence is above the threshold $\lambda_s$, a sequence-wise anomaly is detected in the current window, and the localization mechanism is triggered. As the sliding window is moving along the data stream with stride $\mu$, if no outliers are detected in the previous window, the start position of the sequence-wise anomaly can only exist at the end of $X_t$ in the window $\left \{ x_{t+\omega- \mu},...,x_{t+\omega-1},x_{t+\omega-1} \right \}$ with length $\mu$. Consequently, 
by gradually shifting the \emph{Active window} backward to include one more element in the \emph{Buffer} window (see Fig.~\ref{fig:Srepir}) at a time and calculating the corresponding \emph{DTW} distances as $\left \{ e_1,...,e_i,...,e_{\mu} \right \}$, we can find the maximum $i$ with
$e_i<\lambda_s$, indicating the following element after the \emph{Active window} starting with $i$ is the start of the anomaly subsequence. The \emph{Anomaly flag} is then activated from this position. 
Similarly, to determine the ending position of the anomaly subsequence, we keep sliding the \emph{Active windows} until we find a window with \emph{DTW} distance smaller than $\lambda_s$, indicating that the ending position is within $\left \{ x_{t - \mu},...,x_{t-2},x_{t-1} \right \}$. Again, we shift the \emph{Active window} backwardly one-by-one to include one element of the above window at a time and calculate the corresponding \emph{DTW} distance, until we find the ending position with its \emph{DTW} distance larger than $\lambda_s$.



\section{Experiments}\label{sec:4-exp}
In this section, we conduct extensive experiments to answer the following two questions: 
\emph{Whether DeepFIB outperforms state-of-the-art AD methods} (\textbf{Q1})?~\emph{How does each component of DeepFIB affect its performance} (\textbf{Q2})?


\begin{table*}[h]

\caption{Datasets used in experiments}
\centering
 \resizebox{0.7\textwidth}{!}
{
\begin{tabular}{c|l|c|c|c|c}
\hline
\multicolumn{2}{c|}{Datasets}                 & $\#$Dim & $\#$Train & $\#$Test & Anomaly                      \\ \hline
\multicolumn{2}{c|}{2d-gesture}               & 2           & 8590      & 2420     & 24.63$\%$                       \\ \hline
\multicolumn{2}{c|}{Power demand}                     & 1          & 18145     & 14786    & 11.44$\%$                       \\ \hline
\multirow{6}{*}{ECG} & (A)chfdb\_chf01\_275   & 2           & 2888      & 1772     & 14.61 $\%$                      \\ \cline{2-6} 
                     & (B)chfdb\_chf13\_45590 & 2           & 2439      & 1287     & 12.35 $\%$                      \\ \cline{2-6} 
                     & (C)chfdbchf15          & 2           & 10863     & 3348     & 4.45 $\%$                       \\ \cline{2-6} 
                     & (D)ltstdb\_20221\_43   & 2           & 2610      & 1121     & 11.51 $\%$                      \\ \cline{2-6} 
                     & (E)ltstdb\_20321\_240  & 2           & 2011      & 1447     & 9.61 $\%$                       \\ \cline{2-6} 
                     & (F)mitdb\_100\_180     & 2           & 2943      & 2255     & 8.38 $\%$                       \\ \hline
\multicolumn{2}{c|}{Credit Card}             & 3          & 142403    & 142404   & \multicolumn{1}{c}{0.173 $\%$} \\ \hline
\end{tabular}}
\label{tab:statistics}
\end{table*}

Experiments are conducted on a number of commonly-used benchmark TS datasets, namely \emph{2d-gesture}, \emph{Power demand}, \emph{ECG} and \emph{Credit Card}, ranging from human abnormal behavior detection, power monitoring, healthcare and fraud detection in finance (see Table~\ref{tab:statistics}). As the anomalies in \emph{2d-gesture}, \emph{Power demand}, and \emph{ECG} are mainly sequence outliers, we apply the \emph{DeepFIB-s} model on these datasets. In contrast, the \emph{Credit Card} dataset only contains point outliers, and hence we use \emph{DeepFIB-p} model on it. 

To make a fair comparison with existing models, we use the standard evaluation metrics on the corresponding datasets. For \emph{2d-gesture}, \emph{Power demand} and \emph{Credit Card}, we use precision, recall, and F1-score following~\citep{Shen2020TimeseriesAD}. For \emph{ECG} datasets, we use the AUROC (area under the ROC curve), AUPRC (area under the precision-recall curve) and F1-score, following~\citep{Shen2021TimeSA}. To detect anomalies, we use the maximum anomaly score in each sub-models over the validation dataset to set the threshold. 

More details on experimental settings, additional experimental results and discussions (e.g., hyperparameter analysis) are presented in the supplementary materials.

\subsection{\textbf{Q1}:~Comparison with state-of-the-art methods}

\begin{table*}[h]
\caption{Comparison of anomaly detection performance (as \%), on \emph{2d-gesture} and \emph{Power demand} datasets. The best results are in \textbf{bold} and the second best results are underlined.}
\centering
 \resizebox{\textwidth}{!}
{
\begin{tabular}{c|c|c|c|c|c|c}
\hline
\multirow{2}{*}{Methods} & \multicolumn{3}{c|}{2d-gesture}                                                  & \multicolumn{3}{c}{Power demand}                                                           \\ \cline{2-7} 
                         & precision             & recall              & F1-score                                         & precision              & recall               & F1-score                                          \\ \hline
DAGMM                    & 25.66            & \underline{ \emph{80.47}}            & 38.91                                      & 34.37              & 41.72             & 37.69                                       \\ \hline
EncDec-AD                    & 24.88            & \textbf{100.0}            & 39.85                                      & 13.98              & 54.20             & 22.22                                       \\ \hline
LSTM-VAE                 & 36.62            & 67.72            & 47.54                                      & 8.00             & 56.66             & 14.03                                       \\ \hline
MADGAN                   & 29.41            & 76.4             & 42.47                                      & 13.20             & 60.57              & 27.67                                       \\ \hline
AnoGAN                   & 57.85            & 46.50             & 51.55                                      & 20.28             & 44.41              & 28.85                                       \\ \hline
BeatGAN                  & 55.11            & 45.33            & 49.74                                      & 8.04             & 76.58             & 14.56                                       \\ \hline
OmniAnomaly              & 27.70            & 79.67            & 41.11                                      & 8.55             & \underline{ \emph{78.73}}             & 15.42                                       \\ \hline
MSCRED                   & \underline{ \emph{61.26}}            & 59.11            & 60.17                                      & \underline{ \emph{55.80}}              & 34.32             & 42.50                                       \\ \hline

THOC                     & 54.78            & 75.00            & \underline{ \emph{63.31}}                                      & \textbf{61.50}             & 36.34             & \underline{ \emph{45.68}}                                       \\ \hline

DeepFIB                 & \textbf{93.90 $\pm$ 0.35} & 60.77 $\pm$ 0.24 & \textbf{73.79 $\pm$ 0.19} & 52.21  $\pm$ 0.31 & \textbf{99.99  $\pm$ 0.01} & \textbf{68.60  $\pm$ 0.15} \\ \hline

\end{tabular}}
\begin{tablenotes} 
\tiny
		
		\item - The results of other baselines in the table are extracted from~\citep{Shen2020TimeseriesAD}
 \end{tablenotes} 
\label{tab:2d}
\end{table*}
\textbf{2d-gesture and Power demand}: The results in Table~\ref{tab:2d} show that the proposed \emph{DeepFIB-s} achieves 16.55$\%$ and 50.18$\%$ F1-score improvements on \emph{2d-gesture} and \emph{Power demand}, respectively, compared with the second best methods. 

For \emph{2d-gesture},  the available training data is limited and the temporal relations contained in the data are complex~(body jitter), making it difficult to obtain a discriminative representation in AD models. DAGMM~\citep{Zong2018DeepAG} shows low performance since it does not consider the temporal information of the time-series data at all. 
As for the AD solutions based on generative models  (EncDecAD~\citep{Malhotra2016LSTMbasedEF}, LSTM-VAE~\citep{Park2018AMA}, MAD-GAN~\citep{Li2019MADGANMA}, AnoGAN~\citep{Schlegl2017UnsupervisedAD}, BeatGAN~\citep{Zhou2019BeatGANAR}, OmniAnomaly~\citep{Su2019RobustAD}), they usually require a large amount of training data, limiting their performance in data-scarce scenario. Compared to the above methods, the encoder-decoder architecture MSCRED~\citep{Zhang2019ADN} is relatively easier to train and its AD performance is considerably higher. 
Moreover, the recent THOC~\citep{Shen2020TimeseriesAD} work further improves AD performance by fusing the multi-scale temporal information to capture the complex temporal dynamics. 

The proposed \emph{DeepFIB-s} model outperforms all the above baseline methods since the proposed self-imputation technique allows the model to learn more robust temporal relations from much more self-generated training samples. Notably, we also observe that the \emph{precision} of the \emph{DeepFIB-s} dominates the other baselines.  We attribute it to the anomaly localization algorithm that can locate the anomaly's precise start and end positions, significantly reducing the false positive rate.  


\begin{figure}[h]
    \centering
    \includegraphics[width=0.8\linewidth]{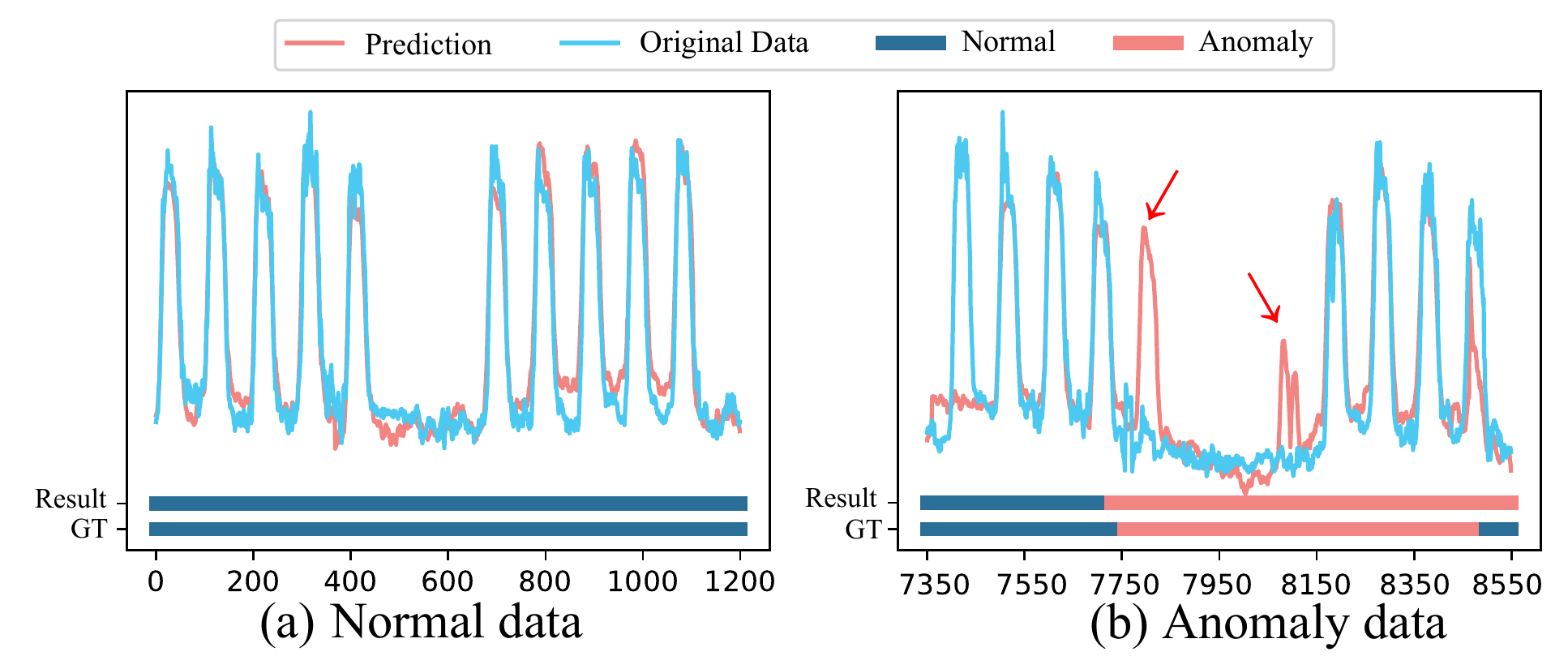}
    \caption{For Power demand dataset, (a) shows two cycles of normal data (0-1200 frame) wherein each cycle contains 5 peaks. (b) shows two cycles with anomaly with missing peaks highlighted using red arrows. The waveform of original data (light blue) is overlaid on the prediction result (light red). 
    Lower color bars show the ground truth (GT) label and our detection result (Result).   
    }
    \label{fig:abnormal}
\end{figure}

For \emph{Power demand}, the data contains many \emph{contextual anomaly\footnote{Contextual anomalies are observations or sequences that deviate from the expected patterns within the time series however if taken in isolation they are within the range of values expected for that signal~\citep{Cook2020AnomalyDF}.}} subsequences (see Fig.~\ref{fig:abnormal}). 
It is quite challenging for existing AD approaches to learn such context information by extracting temporal features from the entire time series as a whole.
In contrast, the proposed sequence-wise masking strategy facilitates learning different kinds of temporal patterns, which is much more effective in detecting such contextual anomalies. As shown in Table~\ref{tab:2d}, the \emph{recall} of our \emph{DeepFIB-s} model almost reaches 100$\%$, indicating all anomalies have been detected. The \emph{precision} is not the best, and we argue that some of the false positives are in fact resulted from the poorly labeled test set (see our supplementary material). 
 \begin{table*}[t]
\caption{Comparison of anomaly detection performance (as \%), on ECG datasets.}
\centering
\resizebox{\textwidth}{!}
{
\begin{tabular}{c|c|c|c|c|c|c|c|c}
\hline
\multirow{2}{*}{Metrics} & \multirow{2}{*}{Methods} & \multicolumn{6}{c|}{ECG}                                                                                                                                                                                                                                                                                                                                                                      & \multirow{2}{*}{Average} \\ \cline{3-8}
                         &                          & A                                                           & B                                                           & C                                                           & D                                                            & E                                                                      & F                                                           &                          \\ \hline
\multirow{6}{*}{AUROC}   & RAE                      & 64.95                                                       & 75.24                                                       & 68.27                                                       & 60.71                                                        & 77.92                                                                  & 44.68                                                       & 65.29                    \\ \cline{2-9} 
                         & RRN                      & 69.50                                                       & 72.07                                                       & 68.49                                                       & 47.05                                                        & 78.81                                                                  & 47.87                                                       & 63.97                    \\ \cline{2-9} 
                         & BeatGAN                  & 66.51                                                       & 73.14                                                       & 58.69                                                       & 59.33                                                        & 82.98                                                                  & 44.19                                                       & 64.14                    \\ \cline{2-9} 
                         & RAE-ensemble             & 68.26                                                       & 77.63                                                       & 70.55                                                       & 64.64                                                        & 83.14                                                                  & 39.66                                                       & 67.31                    \\ \cline{2-9} 
                         & RAMED                    & \underline{ \emph{73.58}} & \underline{ \emph{78.82}} & \underline{ \emph{78.79}} & \underline{ \emph{69.44}}  & \underline{ \emph{83.36}}            & \underline{ \emph{55.64}} & \underline{ \emph{73.27}}                    \\ \cline{2-9} 
                         & DeepFIB                  & \textbf{87.60 $\pm$0.85}                   & \textbf{84.40 $\pm$1.23}                   & \textbf{94.05 $\pm$0.73}                   & \textbf{72.55 $\pm$0.54}                    & \textbf{84.81 $\pm$0.62}                              & \textbf{63.23  $\pm$0.12}                  & \textbf{81.11}                    \\ \hline
\multirow{6}{*}{AUPRC}   & RAE                      & 51.84                                                       & 40.32                                                       & 31.23                                                       & 15.54                                                        & 24.17                                                                  & 7.76                                                        & 28.48                    \\ \cline{2-9} 
                         & RRN                      & 54.90                                                       & 43.13                                                       & 33.49                                                       & 11.63                                                        & 37.68                                                                  & 7.93                                                        & 31.46                    \\ \cline{2-9} 
                         & BeatGAN                  & 52.50                                                       & 44.94                                                       & 19.01                                                       & 14.84                                                        & 34.46                                                                  & 7.66                                                        & 28.90                    \\ \cline{2-9} 
                         & RAE-ensemble             & 56.23                                                       & 54.21                                                       & \underline{ \emph{49.90}} & \underline{ \emph{ 18.47}} & 38.48                                                                  & 7.25                                                        & \underline{ \emph{37.42}}                    \\ \cline{2-9} 
                         & RAMED                    & 56.23                                                       & 54.23                                                       & 34.63                                                       & 17.78                                                        & \textbf{45.78}                                        & \underline{ \emph{10.59}} & 36.54                    \\ \cline{2-9} 
                         & DeepFIB                  & \textbf{85.18$\pm$0.63}                    & \textbf{75.48 $\pm$0.56}                   & \textbf{73.47 $\pm$0.67}                   & \textbf{23.14 $\pm$0.45}                    & \underline{ \emph{38.27 $\pm$ 0.72}} & \textbf{13.16 $\pm$ 0.23}                  & \textbf{51.45}                    \\ \hline
\multirow{6}{*}{F1}      & RAE                      & 52.51                                                       & 49.03                                                       & 32.79                                                       & 25.43                                                        & 33.63                                                                  & 15.47                                                       & 34.81                    \\ \cline{2-9} 
                         & RRN                      & 56.08                                                       & 43.48                                                       & 38.30                                                       & 20.64                                                        & 44.37                                                                  & 15.47                                                       & 36.39                    \\ \cline{2-9} 
                         & BeatGAN                  & 51.93                                                       & 45.18                                                       & 27.99                                                       & 23.67                                                        & 47.02                                                                  & 16.68                                                       & 35.41                    \\ \cline{2-9} 
                         & RAE-ensemble             & \underline{ \emph{56.42}} & \underline{ \emph{52.40}} & \underline{ \emph{58.68}} & 27.75                                                        & 44.98                                                                  & 15.47                                                       & \underline{ \emph{42.62}}                    \\ \cline{2-9} 
                         & RAMED                    & 54.27                                                       & 51.03                                                       & 34.45                                                       & \underline{ \emph{30.87}}  & \underline{ \emph{52.23}}            & \underline{ \emph{20.63}} & 40.58                    \\ \cline{2-9} 
                         & DeepFIB                  & \textbf{80.90 $\pm$0.63}                   & \textbf{78.06 $\pm$0.82}                   & \textbf{78.37 $\pm$0.13}                   & \textbf{44.71 $\pm$0.19}                    & \textbf{58.00 $\pm$0.26}                              & \textbf{34.08 $\pm$0.73}                   & \textbf{62.35}                    \\ \hline
\end{tabular}}
\begin{tablenotes} 
\tiny
		
		\item - The results of other baselines in the table are referred from~\citep{Shen2021TimeSA}
 \end{tablenotes} 
\label{tab:ecg}
\end{table*}

 \textbf{ECG(A-F)}: 
 Compared with (A),(B),(C) datasets, (D),(E),(F) are clearly noisy, which affect the performance of the anomaly detectors significantly. Nevertheless, Table~\ref{tab:ecg} shows that \emph{DeepFIB-s} achieves an average 46.3$\%$ F1-score improvement among all datasets and an impressive 65.2$\%$ improvement for ECG(F) dataset. There are mainly two reasons: (1) the data is scarce~(See Table~\ref{tab:statistics}). Existing AD methods are unable to learn robust temporal relations under such circumstances. In contrast,  the self-imputation training strategy together with the bidirectional forecasting mechanism used in our \emph{DeepFIB-s} model can well address this issue;
 (2) the proposed DTW anomaly score is more effective in detecting the anomaly sequence than the previous point-wise residual scoring~(see Section~\ref{sec:ca} ). Notably, the AUPRC of \emph{DeepFIB} in \emph{ECG(E)} is slightly lower than RAMED~\citep{Shen2021TimeSA}, and we attribute to the fact that some unlabeled sub-sequences are too similar to labeled anomalies in the raw data. 

 \textbf{Credit Card}: Due to the nature of this application, this dataset is stochastic and the temporal relation is not significant.
 Therefore, as shown in Table~\ref{tab:credit}, traditional AD solutions without modeling the underlying temporal dependency achieve fair performance, e.g., OCSVM~\citep{Ma2003TimeseriesND}, ISO Forest~\citep{Liu2008IsolationF}. Besides, the AR~\citep{Rousseeuw1987RobustRA} with a small window size~(e.g., 3, 5) can also identify the local change point without considering longer temporal relations. However, the large \emph{recall} and small \emph{precision} values show its high false positive rates. The prediction-based method, LSTM-RNN~\citep{Bontemps2016CollectiveAD} tries to learn a robust temporal relation from the data, which is infeasible for this dataset. In contrast, the reconstruction-based method, RAE~(recurrent auto-encoder)~\citep{Malhotra2016LSTMbasedEF} performs better since it can estimate the outliers based on the local contextual information.
The proposed \emph{DeepFIB-p} model outperforms all baseline methods, because it can better extract local correlations with the proposed self-imputation strategy. At the same time, compared to our results on other datasets, the relative $26.3\%$ improvement over the second best solution (\emph{AR}) is less impressive and the F1-score with our \emph{DeepFIB-p} model is still less than $25\%$. We attribute it to both the dataset complexity and the lack of temporal relations in this dataset. 
\begin{table*}[h]
\footnotesize
\caption{Comparison of anomaly detection performance (as \%), on Credit Card dataset.}
\centering

\begin{tabular}{c|c|c|c}
\hline
\multirow{2}{*}{Methods} & \multicolumn{3}{c}{Credit Card} \\ \cline{2-4} 
                         & precision  & recall       & F1-score       \\ \hline
AR                       & 11.30     & \textbf{65.20}     & \underline{ \emph{19.20}}    \\ \hline
ISO Forest               & 9.80      & 56.90     & 16.80    \\ \hline
OCSVM                    & 1.70      & \underline{ \emph{62.00}}     & 18.30    \\ \hline
LSTM-RNN                 & 0.40      & 11.00     & 0.70     \\ \hline
RAE                      & \textbf{16.90}     & 21.52     & 18.89    \\ \hline

 \textbf{DeepFIB-p}                & \underline{ \emph{16.52 $\pm$ 0.31}}         & 46.57 $\pm$ 0.41          & \textbf{24.25$\pm$ 0.37}         \\ \hline \hline
 \emph{RAE$^{*}$}                & 13.93 $\pm$ 0.12    & 53.36 $\pm$ 0.21     & 22.07$\pm$ 0.36    \\ \hline
\emph{DeepFIB-p$^\dagger$}       & 16.55 $\pm$ 0.22        & 21.08 $\pm$ 0.12          & 18.50$\pm$ 0.21         \\ \hline
\end{tabular}
\label{tab:credit}
\end{table*}

\subsection{\textbf{Q2}: Ablation study}
In this section, we first evaluate the impact of various components in our \emph{DeepFIB-s} and \emph{DeepFIB-p} models. Next, we replace the SCINet with other sequence models to evaluate its impact.
\subsubsection{Component analysis}\label{sec:ca}
\textbf{DeepFIB-p}:~To demonstrate the impact of the proposed mask-and-impute mechanism in point outlier detection. We add two baseline methods: (1) \emph{DeepFIB-p$^\dagger$}, wherein we remove the self-imputation strategy; (2) \emph{RAE$^*$}, we implement the same mask-and-impute strategy and apply it to the baseline method \emph{RAE}. In Table~\ref{tab:credit}, the performance improvement and degradation of the corresponding variants compared to \emph{DeepFIB-p} and \emph{RAE} clearly demonstrate the effectiveness of the proposed self-imputation strategy for point outlier detection. 

\begin{figure}[h]
    \centering
    \includegraphics[width=0.7\linewidth]{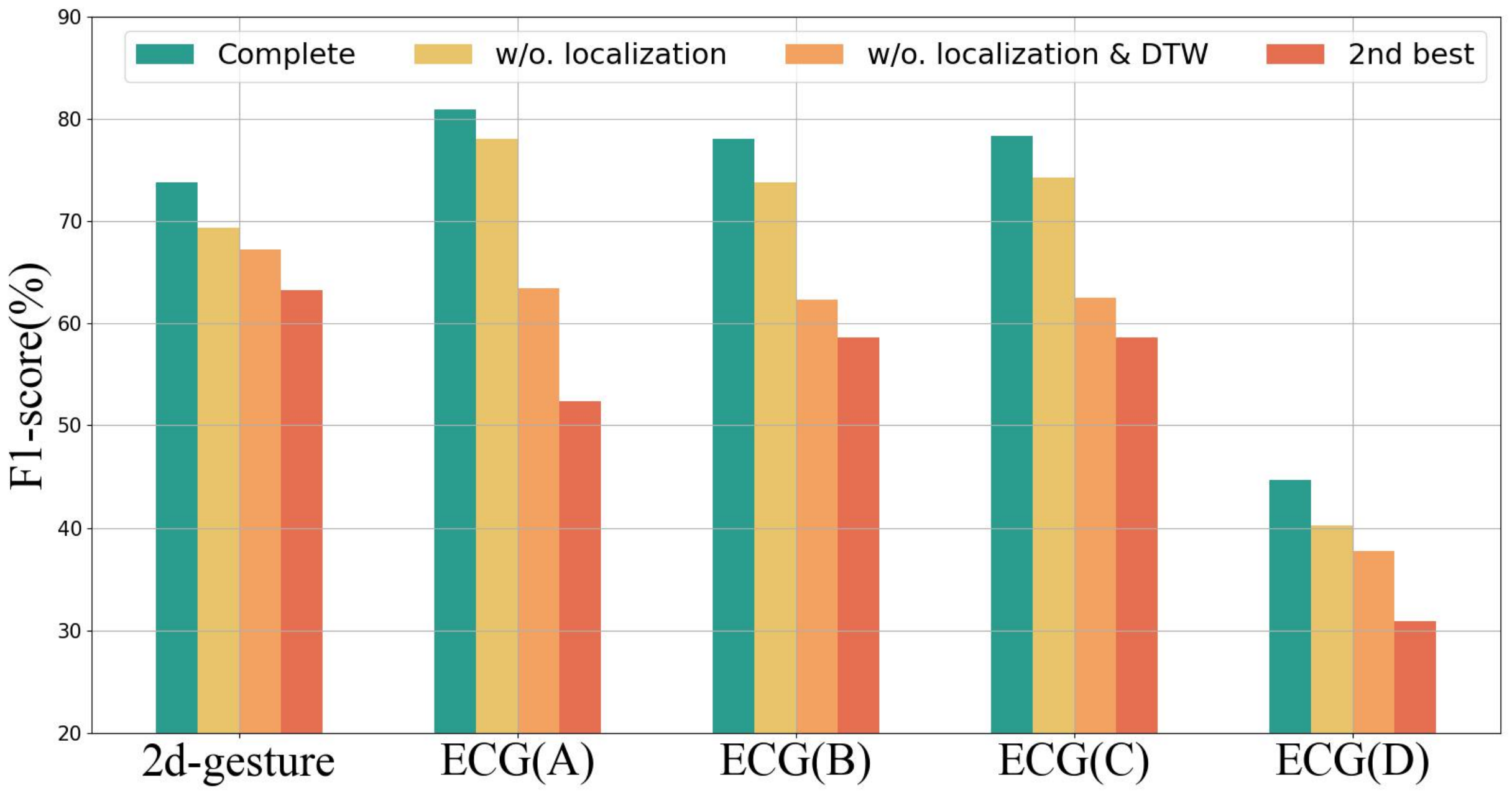}
    \caption{Ablation study of DeepFIB-s for five datasets~(F1-score~$\%$). \emph{2nd best} denotes the previous SOTA results for each dataset.}
    \label{fig:ab}

\end{figure}

\textbf{DeepFIB-s}:~To investigate the impact of different modules of \emph{DeepFIB-s}, we compare two variants of the \emph{DeepFIB-s} on five datasets. The details of the variants are described as below:
For \emph{w/o. localization}, we remove the anomaly localization algorithm from our \emph{DeepFIB-s} model. The \emph{w/o. localization $\&$ DTW} further removes the DTW scoring mechanism, and the anomalies are determined based on point-wise residual errors. As shown in Fig.~\ref{fig:ab}, all these components are essential for achieving high anomaly detection accuracy. At the same time, the proposed self-imputation training strategy is still the main contributor to the performance of our \emph{DeepFIB-s} model, as the results of \emph{w/o. localization $\&$ DTW} are still much better than those of the 2nd best solution. 
Besides, the performance gain of the DTW anomaly scoring indicates that the point-wise outlier estimation is not suitable for evaluating sequence-wise anomalies.

\subsubsection{Impact of SCINet}
In our \emph{DeepFIB} framework, we revise SCINet for time series imputation. To show its impact,
we replace it with other sequence models in \emph{DeepFIB-s}.  As we can see in Table~\ref{tab:sequence}, compared with TCN~\citep{Bai2018AnEE} and LSTM~\citep{Hochreiter1997LongSM}, using SCINet indeed brings significant improvements on F1-sc, whiorech clearly shows its strong temporal relation extraction capability and the effectiveness of the revised architecture for TS imputation. At the same time, compared to the previous SOTA methods (\emph{2nd best}) for the corresponding dataset, with the same mask-and-impute strategy, we can still achieve remarkable performance without using SCINet, indicating the effectiveness of the proposed self-imputation concept itself.  


\begin{table}[h]
\footnotesize
\centering
\caption{The comparison of different sequence models. \emph{2nd best} denotes the previous SOTA methods in each datasets~(~THOC in \emph{2d-gesture} and RAE-ensemble in \emph{ECG(A)}~).}
\begin{tabular}{c|c|c}

\hline
Methods & ECG(A) & 2d-gesture \\ \hline
SCINet  &  \textbf{80.90 $\pm$ 0.63}      &      \textbf{73.79 $\pm$ 0.19}        \\ \hline
TCN     &  69.86  $\pm$ 0.22      &    69.55  $\pm$ 0.28        \\ \hline
LSTM    & 64.16 $\pm$ 0.21      &  66.83  $\pm$ 0.55          \\ \hline
\emph{2nd best}     & 56.42       &  63.31          \\ \hline
\end{tabular}
\label{tab:sequence}
\end{table}




\section{Conclusion}\label{sec:5-conclusion}
In this paper, we propose a novel self-imputation framework \emph{DeepFIB} for time series anomaly detection. Considering the two types of common anomalies in TS data, we implement two mask-and-impute models biased towards them, which facilitate extracting more robust temporal relations than existing AD solutions. Moreover, for sequence-wise anomalies, we propose a novel anomaly localization algorithm that dramatically improves AD detection accuracy. Experiments on various real-world TS datasets demonstrate that DeepFIB outperforms state-of-the-art AD approaches by a large margin, achieving up to more than $65\%$ relative improvement in F1-score.

\bibliography{iclr2022_conference}

\begin{thebibliography}{47}
\providecommand{\natexlab}[1]{#1}
\providecommand{\url}[1]{\texttt{#1}}
\expandafter\ifx\csname urlstyle\endcsname\relax
  \providecommand{\doi}[1]{doi: #1}\else
  \providecommand{\doi}{doi: \begingroup \urlstyle{rm}\Url}\fi

\bibitem[Angiulli \& Pizzuti(2002)Angiulli and Pizzuti]{Angiulli2002FastOD}
F.~Angiulli and C.~Pizzuti.
\newblock Fast outlier detection in high dimensional spaces.
\newblock In \emph{PKDD}, 2002.

\bibitem[Bai et~al.(2018)Bai, Kolter, and Koltun]{Bai2018AnEE}
S.~Bai, J.Z. Kolter, and V.~Koltun.
\newblock An empirical evaluation of generic convolutional and recurrent
  networks for sequence modeling.
\newblock \emph{ArXiv}, abs/1803.01271, 2018.

\bibitem[Bl'azquez-Garc'ia et~al.(2021)Bl'azquez-Garc'ia, Conde, and
  Lozano]{BlazquezGarcia2021ARO}
A.~Bl'azquez-Garc'ia, U.~Conde, A.and~Mori, and J.A. Lozano.
\newblock A review on outlier/anomaly detection in time series data.
\newblock \emph{ACM Computing Surveys (CSUR)}, 54:\penalty0 1 -- 33, 2021.

\bibitem[Bontemps et~al.(2016)Bontemps, Cao, McDermott, and
  Le-Khac]{Bontemps2016CollectiveAD}
L.~Bontemps, V.~L. Cao, J.~McDermott, and N.~A. Le-Khac.
\newblock Collective anomaly detection based on long short-term memory
  recurrent neural networks.
\newblock \emph{International conference on future data and security
  engineering}, abs/1703.09752, 2016.

\bibitem[Cao et~al.(2018)Cao, Wang, Li, Zhou, Li, and Li]{Cao2018BRITSBR}
W.~Cao, D.~Wang, J.~Li, H.~Zhou, L.~Li, and Y.~Li.
\newblock Brits: Bidirectional recurrent imputation for time series.
\newblock \emph{NeurIPS}, abs/1805.10572, 2018.

\bibitem[Chen et~al.(2020)Chen, Kornblith, Norouzi, and Hinton]{Chen2020ASF}
T.~Chen, S.~Kornblith, M.~Norouzi, and G.~Hinton.
\newblock A simple framework for contrastive learning of visual
  representations.
\newblock \emph{PMLR}, abs/2002.05709, 2020.

\bibitem[Chen et~al.(2021)Chen, Chen, Zhang, Yuan, and
  Cheng]{Chen2021LearningGS}
Z.~Chen, D.~Chen, X.~Zhang, Z.~Yuan, and X.~Cheng.
\newblock Learning graph structures with transformer for multivariate time
  series anomaly detection in iot.
\newblock \emph{IEEE Internet of Things Journal}, abs/2104.03466, 2021.

\bibitem[Cook et~al.(2020)Cook, Mısırlı, and Fan]{Cook2020AnomalyDF}
A.~A. Cook, G.~Mısırlı, and Z.~Fan.
\newblock Anomaly detection for iot time-series data: A survey.
\newblock \emph{IEEE Internet of Things Journal}, 7:\penalty0 6481--6494, 2020.

\bibitem[Deldari et~al.(2021)Deldari, Smith, Xue, and Salim]{Deldari2021TimeSC}
S.~Deldari, D.~V. Smith, H.~Xue, and F.~D. Salim.
\newblock Time series change point detection with self-supervised contrastive
  predictive coding.
\newblock \emph{Proceedings of the Web Conference 2021}, 2021.

\bibitem[Deng \& Hooi(2021)Deng and Hooi]{Deng2021GraphNN}
A.~Deng and B.~Hooi.
\newblock Graph neural network-based anomaly detection in multivariate time
  series.
\newblock \emph{ArXiv}, abs/2106.06947, 2021.

\bibitem[Devlin et~al.(2019)Devlin, Chang, Lee, and
  Toutanova]{Devlin2019BERTPO}
J.~Devlin, Ming-Wei Chang, Kenton Lee, and Kristina Toutanova.
\newblock Bert: Pre-training of deep bidirectional transformers for language
  understanding.
\newblock In \emph{NAACL}, 2019.

\bibitem[Falck et~al.(2020)Falck, Sarkar, Roy, and
  Hyland]{Falck2020ContrastiveRL}
F.~Falck, S.K. Sarkar, S.~Roy, and S.L. Hyland.
\newblock Contrastive representation learning for electroencephalogram
  classification.
\newblock 2020.

\bibitem[Fan et~al.(2020)Fan, Zhang, and Gao]{fan2020selfsupervised}
H.~Fan, F.~Zhang, and Y.~Gao.
\newblock Self-supervised time series representation learning by inter-intra
  relational reasoning.
\newblock \emph{arXiv}, abs/2011.13548, 2020.

\bibitem[Fang \& Wang(2020)Fang and Wang]{Fang2020TSImpute}
C.~Fang and C.~Wang.
\newblock {Time Series Data Imputation: A Survey on Deep Learning Approaches}.
\newblock \emph{arXiv}, 2020.

\bibitem[Feng \& Tian(2021)Feng and Tian]{Feng2021TimeSA}
C.~Feng and P.~Tian.
\newblock Time series anomaly detection for cyber-physical systems via neural
  system identification and bayesian filtering.
\newblock \emph{Proceedings of the 27th ACM SIGKDD Conference on Knowledge
  Discovery \& Data Mining}, 2021.

\bibitem[Gupta et~al.(2014)Gupta, Gao, Aggarwal, and Han]{Gupta2014TSADSurvey}
M.~Gupta, J.~Gao, C.C. Aggarwal, and J.~Han.
\newblock {Outlier Detection for Temporal Data: A Survey}.
\newblock \emph{IEEE Transactions on Knowledge and Data Engineering},
  26:\penalty0 2250--2267, 2014.

\bibitem[Hochreiter \& Schmidhuber(1997)Hochreiter and
  Schmidhuber]{Hochreiter1997LongSM}
S.~Hochreiter and J.~Schmidhuber.
\newblock Long short-term memory.
\newblock \emph{Neural Computation}, 9:\penalty0 1735--1780, 1997.

\bibitem[Holt(2004)]{Holt2004ForecastingSA}
C.~Holt.
\newblock Forecasting seasonals and trends by exponentially weighted moving
  averages.
\newblock \emph{International Journal of Forecasting}, 20:\penalty0 5--10,
  2004.

\bibitem[Hu et~al.(2020)Hu, Liu, Gomes, Zitnik, Liang, Pande, and
  Leskovec]{Hu2020StrategiesFP}
W.~Hu, B.~Liu, J.~Gomes, M.~Zitnik, P.~Liang, V.S. Pande, and J.~Leskovec.
\newblock Strategies for pre-training graph neural networks.
\newblock \emph{ICLR}, 2020.

\bibitem[Hundman et~al.(2018)Hundman, Constantinou, Laporte, Colwell, and
  S{\"o}derstr{\"o}m]{Hundman2018DetectingSA}
K.~Hundman, V.~Constantinou, Christopher Laporte, Ian Colwell, and
  T.~S{\"o}derstr{\"o}m.
\newblock Detecting spacecraft anomalies using lstms and nonparametric dynamic
  thresholding.
\newblock \emph{Proceedings of the 24th ACM SIGKDD International Conference on
  Knowledge Discovery \& Data Mining}, 2018.

\bibitem[Keogh et~al.(2005)Keogh, Lin, and Fu]{Keogh2005ECG}
E.~Keogh, J.~Lin, and A.~Fu.
\newblock Hot sax: Efficiently finding the most unusual time series
  subsequence.
\newblock In \emph{ICDM}, pp.\  226--233, 2005.

\bibitem[Kieu et~al.(2019)Kieu, Yang, Guo, and Jensen]{Kieu2019OutlierDF}
T.~Kieu, B.~Yang, C.~Guo, and C.S. Jensen.
\newblock Outlier detection for time series with recurrent autoencoder
  ensembles.
\newblock In \emph{IJCAI}, 2019.

\bibitem[Li et~al.(2019)Li, Chen, Shi, Jin, Goh, and Ng]{Li2019MADGANMA}
D.~Li, D.~Chen, L.~Shi, B.~Jin, J.~Goh, and S.~Ng.
\newblock Mad-gan: Multivariate anomaly detection for time series data with
  generative adversarial networks.
\newblock In \emph{ICANN}, 2019.

\bibitem[Liu et~al.(2008)Liu, Ting, and Zhou]{Liu2008IsolationF}
F.~Liu, K.~Ting, and Z.~Zhou.
\newblock Isolation forest.
\newblock \emph{2008 Eighth IEEE International Conference on Data Mining}, pp.\
   413--422, 2008.

\bibitem[Liu et~al.(2021)Liu, Zeng, and Xu]{Liu2021TimeSI}
M.~Liu, Q.~Zeng, A.and~Lai, and Q~Xu.
\newblock Time series is a special sequence: Forecasting with sample
  convolution and interaction.
\newblock \emph{ArXiv}, abs/2106.09305, 2021.

\bibitem[Liu et~al.(2019)Liu, Yu, Zheng, Zhan, and Yue]{Liu2019NAOMINM}
Y.~Liu, R.~Yu, S.~Zheng, E.~Zhan, and Y.~Yue.
\newblock Naomi: Non-autoregressive multiresolution sequence imputation.
\newblock In \emph{NeurIPS}, 2019.

\bibitem[Luo et~al.(2019)Luo, Zhang, Cai, and Yuan]{Luo2019EGANEG}
Y.~Luo, Y.~Zhang, X.~Cai, and X.~Yuan.
\newblock E²gan: End-to-end generative adversarial network for multivariate
  time series imputation.
\newblock In \emph{IJCAI}, 2019.

\bibitem[Ma \& Perkins(2003)Ma and Perkins]{Ma2003TimeseriesND}
J.~Ma and S.~Perkins.
\newblock Time-series novelty detection using one-class support vector
  machines.
\newblock \emph{Proceedings of the International Joint Conference on Neural
  Networks, 2003.}, 3:\penalty0 1741--1745 vol.3, 2003.

\bibitem[Malhotra et~al.(2016)Malhotra, Ramakrishnan, Anand, Vig, Agarwal, and
  Shroff]{Malhotra2016LSTMbasedEF}
P.~Malhotra, A.~Ramakrishnan, G.~Anand, L.~Vig, P.~Agarwal, and G.M. Shroff.
\newblock Lstm-based encoder-decoder for multi-sensor anomaly detection.
\newblock \emph{ArXiv}, abs/1607.00148, 2016.

\bibitem[Park et~al.(2018)Park, Hoshi, and Kemp]{Park2018AMA}
D.~Park, Y.~Hoshi, and C.C. Kemp.
\newblock A multimodal anomaly detector for robot-assisted feeding using an
  lstm-based variational autoencoder.
\newblock \emph{IEEE Robotics and Automation Letters}, 3:\penalty0 1544--1551,
  2018.

\bibitem[Pathak et~al.(2016)Pathak, Krähenbühl, Donahue, Darrell, and
  Efros]{Pathak2016ContextEF}
D.~Pathak, P.~Krähenbühl, J.~Donahue, T.~Darrell, and A.A. Efros.
\newblock Context encoders: Feature learning by inpainting.
\newblock In \emph{CVPR}, 2016.

\bibitem[Rousseeuw \& Leroy(1987)Rousseeuw and Leroy]{Rousseeuw1987RobustRA}
P.~Rousseeuw and A.~Leroy.
\newblock Robust regression and outlier detection.
\newblock 1987.

\bibitem[Ruff et~al.(2021)Ruff, Kauffmann, Vandermeulen, Montavon, Samek,
  Kloft, Dietterich, and Muller]{Ruff2021ADReview}
L.~Ruff, J.~Kauffmann, R.A. Vandermeulen, G.~Montavon, W.~Samek, M.~Kloft, T.G.
  Dietterich, and K.~Muller.
\newblock {A Unifying Review of Deep and Shallow Anomaly Detection}.
\newblock \emph{Proceedings of the IEEE}, 109\penalty0 (5):\penalty0 756--795,
  2021.
\newblock ISSN 0018-9219.

\bibitem[Saeed et~al.(2021)Saeed, Salim, Ozcelebi, and
  Lukkien]{Saeed2021FederatedSL}
A.~Saeed, F.D. Salim, T.~Ozcelebi, and J.J. Lukkien.
\newblock Federated self-supervised learning of multisensor representations for
  embedded intelligence.
\newblock \emph{IEEE Internet of Things Journal}, 8:\penalty0 1030--1040, 2021.

\bibitem[Sakoe \& Chiba(1978)Sakoe and Chiba]{Sakoe1978DynamicPA}
H.~Sakoe and S.~Chiba.
\newblock Dynamic programming algorithm optimization for spoken word
  recognition.
\newblock \emph{IEEE Transactions on Acoustics, Speech, and Signal Processing},
  26:\penalty0 159--165, 1978.

\bibitem[Schlegl et~al.(2017)Schlegl, Seeböck, Waldstein, S.M.,
  Schmidt-Erfurth, and Langs]{Schlegl2017UnsupervisedAD}
T.~Schlegl, P.~Seeböck, Waldstein, S.M., U.M. Schmidt-Erfurth, and G.~Langs.
\newblock Unsupervised anomaly detection with generative adversarial networks
  to guide marker discovery.
\newblock In \emph{IPMI}, 2017.

\bibitem[Shen \& Kwok(2020)Shen and Kwok]{Shen2020TimeseriesAD}
Z.~Shen, L.and~Li and J.T. Kwok.
\newblock Timeseries anomaly detection using temporal hierarchical one-class
  network.
\newblock In \emph{NeurIPS}, 2020.

\bibitem[Shen et~al.(2021)Shen, Ma, and Kwok]{Shen2021TimeSA}
Z.~Shen, L.and~Yu, Q.~Ma, and J.T. Kwok.
\newblock Time series anomaly detection with multiresolution ensemble decoding.
\newblock In \emph{AAAI}, 2021.

\bibitem[Shyu et~al.(2003)Shyu, Chen, Sarinnapakorn, and Chang]{Shyu2003ANA}
M.~Shyu, Shu-Ching Chen, Kanoksri Sarinnapakorn, and LiWu Chang.
\newblock A novel anomaly detection scheme based on principal component
  classifier.
\newblock 2003.

\bibitem[Su et~al.(2019)Su, Zhao, Niu, Liu, Sun, and Pei]{Su2019RobustAD}
Y.~Su, Y.~Zhao, C.~Niu, R.~Liu, W.~Sun, and D.~Pei.
\newblock Robust anomaly detection for multivariate time series through
  stochastic recurrent neural network.
\newblock \emph{Proceedings of the 25th ACM SIGKDD International Conference on
  Knowledge Discovery \& Data Mining}, 2019.

\bibitem[Yoo et~al.(2021)Yoo, Kim, and Kim]{Yoo2021RecurrentRN}
Y.~Yoo, U.~Kim, and J.~Kim.
\newblock Recurrent reconstructive network for sequential anomaly detection.
\newblock \emph{IEEE Transactions on Cybernetics}, 51:\penalty0 1704--1715,
  2021.

\bibitem[Yu et~al.(2016)Yu, Jibin, and Jiang]{Yu2016AnIA}
Q.~Yu, L.~Jibin, and L.~Jiang.
\newblock An improved arima-based traffic anomaly detection algorithm for
  wireless sensor networks.
\newblock \emph{International Journal of Distributed Sensor Networks}, 12,
  2016.

\bibitem[Zhang et~al.(2019)Zhang, Song, Chen, Feng, Lumezanu, Cheng, Ni, Zong,
  Chen, and Chawla]{Zhang2019ADN}
C.~Zhang, D.~Song, Y.~Chen, X.~Feng, C.~Lumezanu, W.~Cheng, J.~Ni, B.~Zong,
  H.~Chen, and N.a Chawla.
\newblock A deep neural network for unsupervised anomaly detection and
  diagnosis in multivariate time series data.
\newblock \emph{AAAI}, abs/1811.08055, 2019.

\bibitem[Zhang et~al.(2016)Zhang, Isola, and Efros]{Zhang2016ColorfulIC}
R.~Zhang, P.~Isola, and A.A. Efros.
\newblock Colorful image colorization.
\newblock In \emph{ECCV}, 2016.

\bibitem[Zheng et~al.(2018)Zheng, Zhou, Sheng, Xue, and
  Chen]{Zheng2018GenerativeAN}
Y.~Zheng, X.~Zhou, W.~Sheng, Y.~Xue, and S.~Chen.
\newblock Generative adversarial network based telecom fraud detection at the
  receiving bank.
\newblock \emph{Neural networks : the official journal of the International
  Neural Network Society}, 102:\penalty0 78--86, 2018.

\bibitem[Zhou et~al.(2019)Zhou, Liu, Hooi, Cheng, and Ye]{Zhou2019BeatGANAR}
B.~Zhou, S.~Liu, B.~Hooi, X.~Cheng, and J.~Ye.
\newblock Beatgan: Anomalous rhythm detection using adversarially generated
  time series.
\newblock In \emph{IJCAI}, 2019.

\bibitem[Zong et~al.(2018)Zong, Song, Min, Cheng, Lumezanu, Cho, and
  Chen]{Zong2018DeepAG}
B.~Zong, Q.~Song, M.~Min, W.~Cheng, C.~Lumezanu, D.~Cho, and H.~Chen.
\newblock Deep autoencoding gaussian mixture model for unsupervised anomaly
  detection.
\newblock In \emph{ICLR}, 2018.

\end{thebibliography}
\bibliographystyle{iclr2022_conference}


\end{document}